\documentclass[sigconf]{acmart}
\usepackage{graphicx}
\usepackage{subcaption}
\usepackage[many]{tcolorbox}
\usepackage{tabularx}
\usepackage{booktabs}
\usepackage{hyperref}
\usepackage{makecell}
\usepackage{float}

\usepackage{tikz}
\usepackage{standalone}
\usepackage{pgf-pie}

\usepackage{mdframed}

\usepackage{pgfplots}
\pgfplotsset{compat=newest}
\usepgfplotslibrary{groupplots}
\usetikzlibrary{pgfplots.statistics}

\usepackage{xparse}   
\NewDocumentCommand{\rot}{O{90} O{1em} m}{\makebox[#2][l]{\rotatebox{#1}{#3}}}%

\usepackage{amsthm}
\usepackage{amsmath,amsfonts}

\usepackage{xspace}

\newcommand{\inn}[1]{\scalebox{0.9}{\colorbox{amethyst!30}{#1}}}
\newcommand{\out}[1]{\scalebox{0.9}{\colorbox{deepteal!30}{#1}}}

\definecolor{mongo}{RGB}{236, 142, 51}

\definecolor{royalblue}{RGB}{65,102,245}

\DeclareMathAlphabet{\mathsl}{OT1}{ptm}{m}{sl}

\usepackage{xcolor}
\usepackage{csvsimple}
\usepackage{siunitx}
\usepackage{booktabs}
\usepackage[np]{numprint}
\usepackage{multirow}

\usepackage[inline]{enumitem}
\setlist[description]{leftmargin=1em}
\setlist[itemize]{leftmargin=1em}
\setlist[enumerate]{leftmargin=1.5em}
\usepackage{pifont}
\usepackage{tikz}
\usepackage{standalone}
\usepackage{tikz-qtree}
\usepackage{float}
\pgfplotsset{ every non boxed x axis/.append style={x axis line style=-},
     every non boxed y axis/.append style={y axis line style=-}}

\definecolor{amethyst}{rgb}{0.6, 0.4, 0.8}
\definecolor{deepteal}{RGB}{0, 128, 128}

\usepackage[normalem]{ulem}

\setcopyright{acmlicensed}
\copyrightyear{2018}
\acmYear{2018}
\acmDOI{XXXXXXX.XXXXXXX}

\acmConference[Under review]{Under review, 2024}
\acmISBN{978-1-4503-XXXX-X/18/06}

\begin{document}

\title{Reasoner Outperforms: Generative Stance Detection with Rationalization for Social Media}

\author{Jiaqing Yuan}
\affiliation{%
  \institution{North Carolina State University}
  \city{Raleigh}
  \state{North Carolina}
  \country{USA}
}
\email{jordanyuan111@gmail.com}

\author{Ruijie Xi}
\affiliation{%
  \institution{North Carolina State University}
  \city{Raleigh}
  \state{North Carolina}
  \country{USA}
}
\orcid{0000-0003-0941-5088}
\email{rxi@ncsu.edu}

\author{Munindar P. Singh}
\affiliation{%
  \institution{North Carolina State University}
  \city{Raleigh}
  \state{North Carolina}
  \country{USA}
}
\orcid{0000-0003-3599-3893}
\email{mpsingh@ncsu.edu}

\begin{abstract}

Stance detection is crucial for fostering a human-centric Web by analyzing user-generated content to identify biases and harmful narratives that undermine trust. 
With the development of Large Language Models (LLMs), existing approaches treat stance detection as a classification problem, providing robust methodologies for modeling complex group interactions and advancing capabilities in natural language tasks. 
However, these methods often lack interpretability, limiting their ability to offer transparent and understandable justifications for predictions.
This study adopts a generative approach, where stance predictions include explicit, interpretable rationales, and integrates them into smaller language models through single-task and multitask learning.
We find that incorporating reasoning into stance detection enables the smaller model (FlanT5) to outperform GPT-3.5's zero-shot performance, achieving an improvement of up to 9.57\%.
Moreover, our results show that reasoning capabilities enhance multitask learning performance but may reduce effectiveness in single-task settings. 
Crucially, we demonstrate that faithful rationales improve rationale distillation into SLMs, advancing efforts to build interpretable, trustworthy systems for addressing discrimination, fostering trust, and promoting equitable engagement on social media.

\end{abstract}


%


\maketitle

\section{Introduction}

Stance detection plays a crucial role in fostering a human-centric Web by promoting fairness, inclusivity, and accountability in online interactions. 
By analyzing user-generated content, stance detection identifies biases, discriminatory language, and harmful narratives, which can undermine trust and foster division \citep{de-2024-decipher}. 
The Web Science community has extensively explored stance detection and related topics, addressing critical areas such as political expression \citep{graells-2020-stance}. 
These efforts are instrumental in strengthening Web governance, fostering trust, and cultivating healthier online ecosystems across platforms \citep{li-2024-pro, aldayel-2021-stance}.
Stance detection involves classifying an author's opinion (\emph{in favor of}, \emph{against}, or \emph{neutral}) regarding specific subjects, such as topics or claims \citep{10.5555/3455716.3455856, ng-2022-my}. Existing approaches predominantly treat stance detection as a classification problem \citep{allaway-etal-2021-adversarial, yang-urbani-2021-tribrid, xu-etal-2022-openstance, de-2024-decipher, graells-2022-bots, zhang-2023-wear, li-2024-pro}, often prioritizing accuracy but lacking interpretability.

Understanding the reasoning behind stances is crucial for AI systems to effectively interpret nuanced opinions and foster a more equitable and ethical online environment. 
Analyzing the structural and linguistic properties of stance reasoning is key to uncovering insights into opinion dynamics, combating misinformation, and addressing the evolution of harmful behaviors \citep{de-2024-decipher, graells-2022-bots, zhang-2023-wear}.
Generating human-labeled rationales for stance detection is costly and requires domain-specific expertise \citep{10.5555/3327546.3327624}. 
Large language models (LLMs), particularly those with Chain-of-Thought (CoT) reasoning capabilities \citep{Wei2022ChainOT}, have shown exceptional performance in complex tasks like multihop question answering \citep{NEURIPS2022_11332b6b} and mathematical problem-solving \citep{Wei2022ChainOT}. Moreover, machine-generated instruction-following data has enabled LLMs to achieve remarkable zero-shot capabilities across diverse tasks \citep{wang-2022-super, peng-2023-instruction}. 
While prior work has demonstrated that GPT's explanations can improve the interpretability of stance detection predictions \citep{zhang-2022-would}, no studies have yet leveraged these explanations for further research or practical applications in stance detection.
\begin{figure}[!htb]
\centering
\includegraphics[clip, trim=0cm 6cm 16cm 2cm, scale=0.65]{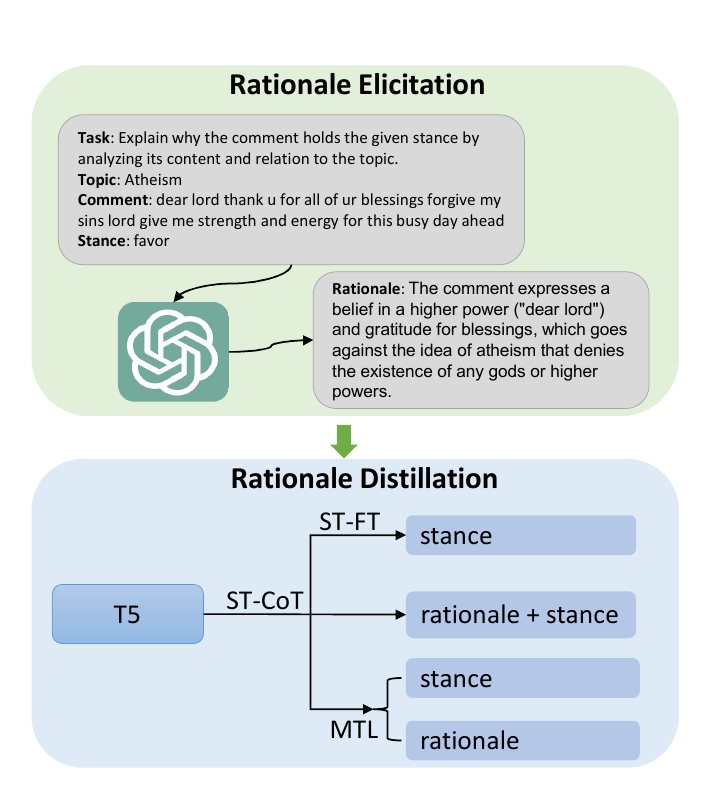}
\Description{place holder}
\caption{Framework for an explainable stance detection system. GPT-3.5 generates rationales conditioned on ground truth labels, and rationale distillation compares (a) ST-FT, (b) ST-CoT, and (c) MTL, where stance labels and rationales are generated concurrently.}
\label{fig:pipeline}
\end{figure}

To address the research gaps, we propose an explainable framework leveraging generative models to produce both predictions and rationales. 
Specifically, we use GPT-3.5 to generate CoT predictions for stance detection. 
As shown in Figure~\ref{fig:pipeline}, direct CoT prompting often produces unfaithful rationales, which we address by conditioning rationale generation on ground-truth labels to improve coherence.
We evaluate two rationale distillation paradigms for small language models (SLMs): single-task CoT (ST-CoT) and multitask learning (MTL), comparing them to single-task finetuning (ST-FT). ST-CoT generates rationales before predictions, while MTL jointly generates both, offering greater flexibility.

Our results show that MTL outperforms other methods, particularly in low-data scenarios, highlighting its robustness for diverse Web applications. 
Instruction tuning further reveals that rationale distillation benefits SLMs more than instruction-tuned models. 
Our contributions include (1) methods for eliciting faithful rationales from LLMs, (2) demonstrating MTL's superiority for rationale distillation, and (3) identifying key factors driving rationale-based learning in stance detection.

\section{Related Work}
Early stance detection relied on machine learning models with handcrafted features \citep{Dey2017TwitterSD, Aldayel2019YourSI}. Subsequent research adopted deep learning approaches, including recurrent networks \citep{kochkina-etal-2017-turing}, attention mechanisms \citep{10.1016/j.ipm.2019.03.010}, and pretrained models like BERT \citep{Devlin2019BERTPO}, driving performance improvements. Chain-of-thought (CoT) prompting elicits reasoning in complex tasks, with \citet{Wei2022ChainOT} demonstrating step-by-step reasoning using few-shot learning and \citet{Kojima2022LargeLM} introducing zero-shot-CoT via the prompt ``Let's think step by step,'' achieving state-of-the-art performance in tasks like arithmetic and logical reasoning. 

\section{Methodology}
Figure~\ref{fig:pipeline} illustrates the framework for training a rationalized stance detection system. Human annotation for stance rationales is resource-intensive and prone to inconsistencies due to varying annotator expertise. To address this, we leverage the CoT capability of LLMs for consistent and faithful rationale generation. For integrating rationales into SLM training, we compare two approaches: single-task chain-of-thought (ST-CoT), which enforces rationale generation before prediction, and multitask learning (MTL), which separates the tasks. ST-CoT enforces rationale generation before prediction, intuitively appearing more effective than MTL, where prediction and rationale generation are separate. However, our experimental results contradict this expectation, showing MTL performs better.

\begin{table}[!htb]
\centering
\begin{tabular}{l r r r}
\toprule
 Topic & Favor & Against & Neutral\\ 
 \midrule
    Donald Trump & 148 & 299 & 260 \\ 
    Hillary Clinton & 163 & 565 & 356\\ 
    Feminist Movement & 268 & 511 & 170\\ 
    Legalization of Abortion & 167 & 544 & 222\\ 
    Atheism & 124 & 464 & 145\\ 
    Climate Change is Concern & 335 & 26 & 203\\ 
 \bottomrule
\end{tabular}
\caption{Statistics of SemEval-2016.}
\label{table:sem16}
\end{table}

\subsection{Elicitation of Rationales from LLMs}
\label{sec:elicitation}

Box~\ref{box:prompts} presents the prompts used in this study, with colored text distinguishing input (\inn{input}) and output (\out{output}). 
Our initial approach used GPT-3.5 to classify stances using Prompt 1. 
However, similar to previous work \citep{zhang-2022-would}, its performance on SemEval-2016 is modest (average F1 = 70.15\% across three random runs) given its scale, highlighting the challenges of directly applying GPT-3.5 to stance detection and the need for fine-tuning a custom model.
To obtain accurate rationales, we reframe the predictive task as an explanatory one using Prompt 2 and GPT-3.5, which has been shown to generate robust explanations for stance detection \citep{zhang-2022-would}.
Conditioning rationales on ground-truth labels encouraged GPT-3.5 to establish connections between the comment and topic. 
While prior work on rationale extraction often relies on in-context learning with human-written exemplars \citep{Wei2022ChainOT}, we adopt a zero-shot approach to eliminate human effort. 
Specifically, we prompt the model with \emph{The comment is classified as [Stance] towards [Topic] because}, regulating the consistency between the rationale and the ground-truth label.

\subsection{Rationale Distillation into SLM}

We now present three finetuning paradigms designed to facilitate the adoption of a generative approach for the task of stance classification. Our chosen foundation model for finetuning is the unified text-to-text encoder-decoder model known as T5 \citep{10.5555/3455716.3455856}. We choose T5 for its unified text-to-text encoder-decoder architecture, well-suited for generating predictions and rationales, and its strong performance across diverse NLP tasks. We now introduce the details of our approaches.

\begin{description}
    \item[Single-task regular finetuning (ST-FT):] The conventional approach in leveraging generative models for classification tasks involves a verbalizer \citep{mosbach-etal-2023-shot}, a mechanism that associates each numerical label with a corresponding word or a set of words. Here, we employ the terms \emph{favor, against, neutral} as the definitive labels representing the stance in our model's ground truth. This methodology aligns with the established convention of transforming numerical outputs into interpretable linguistic expressions, facilitating a more intuitive comprehension of the generated classifications. ST-FT excludes the reasoning component during prediction.
    \item[Single-task chain-of-thought finetuning (ST-CoT):] An emerging capacity within LLMs is to articulate intermediate reasoning steps in the resolution of intricate problems prior to arriving at a conclusive solution. In contrast, SLMs exhibit a deficiency in respect of such proficiency. Therefore, an approach to equip SLMs with this capability involves leveraging rationale as additional finetuning supervision. The training process requires SLMs to generate rationales before making predictions, compelling the model to capture the relationship between rationales and predictions.
    \item[Multitask learning (MTL):] ST-CoT enforces rationale generation before predictions, but its rigidity can hinder performance with sparse training data, as poor rationale quality often leads to incorrect predictions. To address this, we propose a multitask learning (MTL) approach that treats reasoning as a flexible supervisory signal. MTL allows simultaneous generation of predictions and rationales, enhancing both robustness and interpretability. The MTL loss is defined as $\mathcal{L} = \alpha\mathcal{L}_{\text{stance}} + (1-\alpha)\mathcal{L}_{\text{rationale}}$, 
    where \(\alpha\) controls the weight of each task.
    For both predictions, we use the cross-entropy loss. Rationale prediction uses a token-level loss, treating generation as a causal language modeling task, with the model predicting tokens relative to the rationale sequences. Unlike classification models such as BERT \citep{devlin-etal-2019-bert}, where separate classification heads are added for specific tasks, we adopt the multitask learning paradigm of T5 \citep{10.5555/3455716.3455856}. Specifically, we prepend task-specific prefixes to each input: \emph{Stance:} for the stance prediction task and \emph{Explain:} for the rationale generation task.
 
\end{description}

\begin{tcolorbox}[
    colback=black!5!white,
    colframe=gray!150!black,
    title=Prompts,
    fonttitle=\bfseries,
    sharp corners,
    boxrule=0.5pt, 
    left=1mm ,
    right=1mm,
    top=1mm,
    bottom=1mm
]
\label{box:prompts}
\textbf{Prompt 1}  
Your task is to classify the stance of the comment on the topic as ``favor'', ``against'', or ``neutral''. Conclude with the label.
\\
\inn{Topic: \{Topic\}}\\
\inn{Comment: \{Comment\}}  \\
\out{Stance:} 

\textbf{Prompt 2}  
Explain the stance of the comment towards the topic by analyzing its content and relation to the topic.

Your answer should be begin with: ``The \inn{Comment: \{Comment\}} is classified as \out{Stance:} towards \inn{Topic: \{Topic\}} because''  


\textbf{Prompt 3}  
Your task if to classify the stance of the comment on the topic as ``favor'', ``against'', or ``neutral''.  
\\
\inn{Topic: \{Topic\}}\\
\inn{Comment: \{Comment\}}  \\
\out{Stance:}  \\
\out{Explain:} 
\end{tcolorbox}

\section{Experiments}
\paragraph{Dataset}
We evaluate our framework on the SemEval-2016 Task 6 Subtask \citep{mohammad-etal-2016-semeval}, containing \np{4163} English tweets annotated as favor, against, or neutral across five topics: Atheism (AT), Climate Change (CC), Feminist Movement (FM), Hillary Clinton (HC), and Legalization of Abortion (LA). 
Table~\ref{table:sem16} displays the statistics of SemEval-2016.
Following prior work, we use the official train-test split but create a new validation set by randomly sampling 10\% of the training data, ensuring all topics are represented.

\paragraph{Evaluation Metric}
We use the macro-average F1-score for \emph{favor} and \emph{against} ($F_{avg} = \frac{F_{favor} + F_{against}}{2}$) as the evaluation metric, with \emph{neutral} included in training and testing.

\paragraph{Rationale Distillation}
We finetune T5 \citep{10.5555/3455716.3455856} and its instruction-tuned counterpart FlanT5 \citep{Chung2022ScalingIL} at small (80M), base (250M), and large (780M) sizes. For ST-FT and ST-CoT, inputs are the target and comment separated by an end-of-sentence token. For MTL, inputs are prefixed with \emph{Stance:} or \emph{Explanation:} depending on the task. FlanT5 uses a unified input format, shown in Prompt 3.

\paragraph{Experimental Details}
We train models with a batch size of 128, learning rate 5e-5, 30 epochs, maximum input length 512, and maximum generation length 256, using NVIDIA GPUs (A100, A30, A10, A6000).

\begin{table}[!htb]
\centering
\begin{tabular}{l l r r}
\toprule
 Size & Task & T5 & FlanT5\\ 
 \midrule
 \multirow{3}{*}{\shortstack{Small \\ (80M)}} & ST-FT & $60.81_{5.91}$ & $66.30_{0.19}$\\ 
& ST-CoT & $50.05_{0.32}$ & $54.13_{0.28}$\\ 
& MTL & $64.66_{4.16}$ & $66.80_{0.18}$\\ 
 \midrule

\multirow{3}{*}{\shortstack{Base \\ (250M)}} & ST-FT & $65.54_{3.97}$ & $73.56_{0.30}$\\ 
 & ST-CoT & $58.46_{0.73}$ & $60.89_{0.37}$\\ 
 & MTL & $67.40_{1.11}$ &  $74.53_{0.34}$\\ 
 \midrule
 
\multirow{3}{*}{\shortstack{Large \\ (780M)}} & ST-FT & $68.46_{0.45}$ & $78.76_{0.29}$\\ 
 & ST-CoT & $64.47_{0.28}$ & $72.29_{0.36}$\\ 
 & MTL & $76.79_{0.71}$ & $79.72_{0.23}$\\ 

 \bottomrule
\end{tabular}
\caption{$F_{avg}$ scores across models and sizes, with MTL performance reported at the optimal $\alpha$. The subscripts are the standard deviation across three random runs. We establish the baseline by using GPT-3.5 to classify stances with Prompt 1, achieving an F1 score of 70.15\% across three random runs.}
\label{table:results}
\end{table}

\begin{figure*}[!htb]
\centering
\includegraphics[scale=0.37]{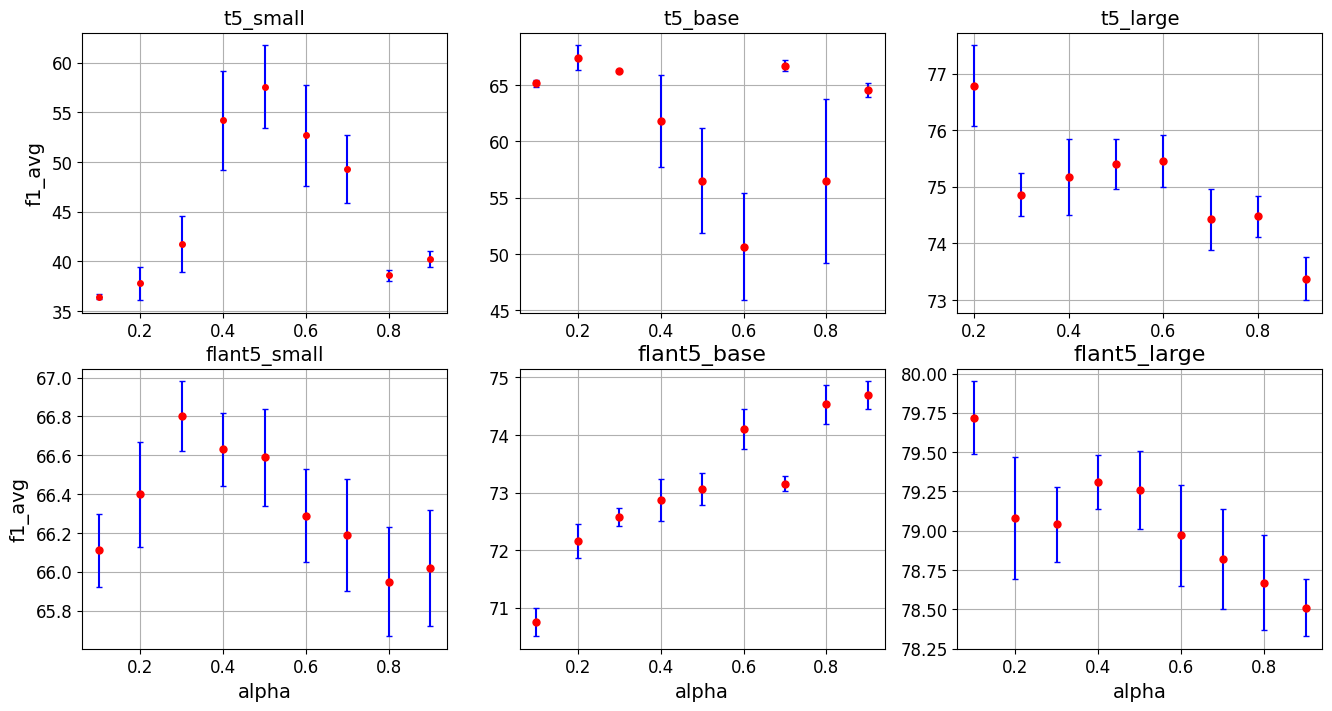}
\Description{place holder}
\caption{$F_{avg}$ scores on different $\alpha$ across models. Standard deviations are calculated from three random runs.}
\label{fig:alpha}
\end{figure*}

\begin{figure*}[!htb]
\centering
\includegraphics[scale=0.37]{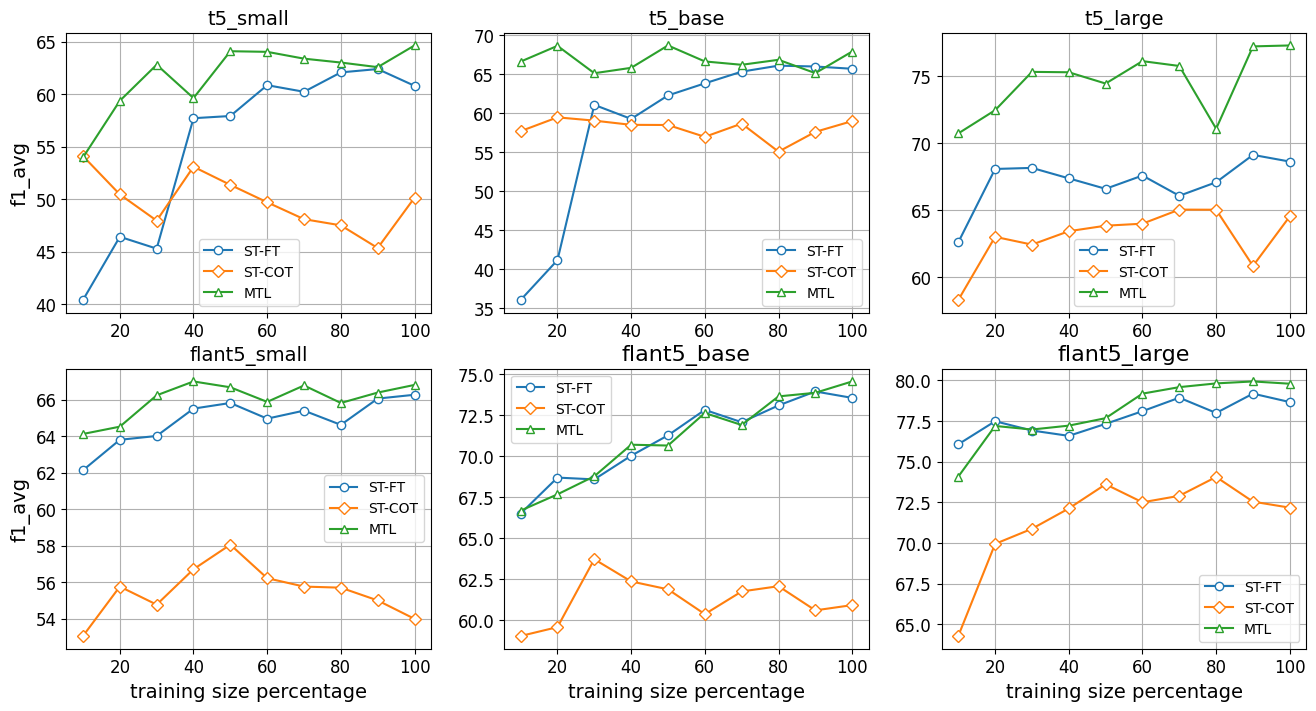}
\Description{place holder}
\caption{$F_{avg}$ for training with different sizes, ranging from 10\% to 100\%. }
\label{fig:size}
\end{figure*}

\section{Results}
Table~\ref{table:results} shows $F_{avg}$ across tasks and models. Compared to the zero-shot baseline performance of 70.15\% using GPT-3.5 with Prompt 1 (as described in Section~\ref{sec:elicitation}), our approaches achieve up to 79.72\%, even with the relatively small size of T5 models. We summarize our findings below. 

\begin{description}
\item[MTL enhances ST performance across all settings:] MTL consistently outperforms ST-FT by independently generating rationales, avoiding interference with prediction tasks. This fosters deeper rationale-prediction connections. T5 models benefit more from MTL than FlanT5, with average improvements of 4.71 and 0.81, respectively, indicating that instruction tuning equips FlanT5 with inherent reasoning capabilities, reducing the marginal effects of further finetuning.
\item[SLMs struggle with CoT capabilities:] ST-CoT underperforms compared to ST-FT across all settings, as generating intermediate steps is challenging for SLMs with limited data (around \np{3000} examples). Deviations in rationale generation often lead to failed or incorrect predictions, making rationale as an auxiliary task a more effective approach.
\item[Instruction tuning improves performance:] FlanT5, despite its smaller size, rivals or outperforms larger T5 models. For instance, FlanT5-Small achieves 66.30 in ST-FT, surpassing T5-Base's 65.54, and FlanT5-Base scores 74.53 in MTL, compared to T5-Large's 76.79. FlanT5's finetuning on diverse datasets enhances its instruction-following proficiency, improving stance classification performance.
\item[Weighting rationale generation varies performance:] We examine the impact of the prediction and rationale generation tasks in MTL by varying the parameter $\alpha$ from 0.1 to 0.9, as shown in Figure~\ref{fig:alpha}. Optimal $\alpha$ values are generally low: 0.5 for T5-Small, 0.2 for T5-Base and T5-Large, 0.3 for FlanT5-Small, and 0.1 for FlanT5-Large, with FlanT5-Base as an exception at 0.9. These results suggest that rationale distillation enhances MTL performance, likely due to the stronger supervision signal provided by the more challenging rationale generation task.
\item[Training data size affect performance:] We evaluate generative models under limited training data by varying dataset size from 10\% to 100\% (Figure~\ref{fig:size}). FlanT5 and T5 models show consistent performance improvements with larger training sizes for ST-FT and MTL. However, ST-CoT exhibits greater variability, indicating potential instability during training. Notably, T5-Base and T5-Large achieve performance comparable to full-data training with only 10-20\% of the dataset under MTL, outperforming ST-FT and ST-CoT. 
\end{description}

\section{Conclusions}

Effective stance detection allows AI systems to align with societal values, promoting a Web where ethical considerations guide user interactions and supporting a human-centric approach to technology.
In this paper, we present an explainable stance detection system that leverages generative models to produce both predictions and justifications, reframing the task with GPT-3.5 to prioritize rationale clarity, given its strong performance in generating explanations for stance detection \citep{zhang-2022-would}.
By providing interpretable justifications, our method enhances the reliability of automated systems in detecting biases, misinformation, and polarizing narratives, thereby fostering a safer and more inclusive online environment. 
We propose two rationale distillation methods--ST-CoT and MTL--demonstrating that MTL outperforms, particularly in low-data settings. 
Additionally, small language models benefit more from justification distillation than instruction-tuned models, emphasizing the importance of both prediction and reasoning in MTL. 
This work contributes to advancing stance detection by promoting informed user engagement, supporting governance against harmful content, and aiding in bridging digital divides through improved interpretability and accessibility of web technologies. Future research extending these methods to larger, cross-dataset applications \citep{ng-2022-my} could further enhance scalable solutions for fostering trust and inclusivity on the web.

\bibliographystyle{ACM-Reference-Format}
\bibliography{Jiaqing, Ruijie}
\end{document}